# Incremental Risk Assessment of Progressive Elder Financial Scams via Instruction-Tuned Small Language Models

Parviz Ghafariasl[1], Weimin Fu[2], Xiaolong Guo[2]*, Shing I. Chang[1]*

[1]*Department of Industrial and Manufacturing Systems Engineering*

[2]*Department of Electrical and Computer Engineering*

*Kansas State University, Manhattan, KS, USA*

Email: {parvizghafari, weiminf, guoxiaolong, changs}@ksu.edu

***Abstract***

Financial scams targeting older adults increasingly occur through text and voice channels such as email, SMS, and phone calls, unfolding over multiple conversational turns that begin with impersonation or casual contact, escalate through trust building and urgency, and culminate in requests for sensitive information or financial transfers. Because risk signals emerge incrementally across turns, effective detection requires models that continuously update risk estimates under resource-constrained deployment settings. We propose a cumulative turn-based risk assessment framework that incrementally aggregates conversational turns and re-estimates risk at each step, enabling dynamic scam monitoring across progressively evolving conversations. A multi-turn dialogue dataset is constructed to cover investment, charity, and tech support scam scenarios, with each dialogue containing two to eight turns and annotated at every cumulative stage with a qualitative risk level, a continuous risk score, an explanatory rationale, and a safety recommendation. Four small language models (Phi-4, LLaMA-3.2, DeepSeek-R1, and Qwen3) are fine-tuned and evaluated under a unified training framework. Fine-tuned small models capture fraud-related linguistic cues and cross-turn escalation patterns while maintaining compact architectures suitable for mobile and resource-constrained deployment settings. Among the evaluated models, Phi-4 and LLaMA-3.2 achieve stronger turn-aware risk estimation performance relative to their parameter scale. These results suggest that structured cumulative modeling can support incremental scam risk assessment in deployment-oriented settings while highlighting the potential of compact language models for privacy-aware and on-device fraud protection.

## I. Introduction

Financial scams targeting older adults have increased in recent years, posing risks to financial security and public trust. In 2024, the FBI Internet Crime Complaint Center reported that cyber-enabled fraud accounted for approximately 84% of total financial losses in the United States, with adults aged 60 and older submitting over 147,000 complaints and reporting an estimated $4.8 billion in losses (FBI IC3, 2024). Preliminary data from the Federal Trade Commission indicate that older adults reported more than $745 million in scam-related losses during the first quarter of 2025 alone (Federal Trade Commission, 2025). These statistics reflect both the scale and persistence of financial exploitation among older populations.

Scams unfold through unsolicited digital communications such as email, SMS, or phone calls, often exploiting social isolation, trust, and age-related vulnerabilities (AARP, 2026). Rather than occurring in a single message, scam attempts typically escalate across multiple conversational turns, progressing from benign or impersonation-based contact to trust building, urgency framing, and psychological coercion before culminating in financial harm (Wood & Lichtenberg, 2017). Effective prevention therefore depends on detecting risk incrementally as conversations evolve.

Detecting such progressive and context-dependent patterns remains challenging for conversational natural language processing systems. While large language models demonstrate strong reasoning capabilities, their computational demands, deployment complexity, and privacy implications limit their suitability for real-time, on-device elder protection. Existing approaches to fraud detection often rely on retrospective analysis of complete conversations and computationally intensive models, and treat dialogues as static samples rather than temporally evolving interactions (Ali & Ghanem, 2025). These limitations hinder deployment in mobile and resource-constrained environments. To address these challenges, this study proposes a structured, cumulative turn-based risk assessment framework for elderly scam detection that operates on-device using small language models. Instead of treating dialogues as static text inputs, conversational turns are incrementally aggregated and risk is re-estimated at each step, enabling real-time monitoring within mobile resource constraints. A balanced synthetic multi-turn dialogue dataset spanning investment- and charity-related scams and tech support–related scams is constructed to support training and evaluation. Through a unified experimental framework, compact instruction-tuned small language models are fine-tuned and assessed for turn-aware risk estimation under deployment constraints.

Our contributions are

- We introduce a cumulative turn-based risk modeling framework that supports incremental and real-time scam detection for older adults.
- We construct a balanced multi-turn dialogue dataset with fine-grained turn-level supervision and cumulative dialogue representation to model temporal risk escalation.
- We demonstrate that compact instruction-tuned small language models achieve competitive risk estimation performance while maintaining model scales suitable for deployment-oriented and resource-constrained settings.

## II. Background

LLMs have achieved success in task understanding and reasoning across diverse applications. Nevertheless, their high computational cost, large memory footprint, and deployment complexity restrict their suitability for real-time and on-device use, particularly in safety-critical scenarios. SLMs mitigate these limitations by adopting compact architectures—typically comprising hundreds of millions to a few billion parameters—while preserving sufficient semantic expressiveness and reasoning capability for domain-specific tasks. Recent studies indicate that SLMs are especially well suited for mobile and edge environments, where constraints on latency, energy consumption, and data privacy are paramount (Xu et al., 2024). Rather than relying on scale alone, SLMs prioritize efficient architectural design and targeted adaptation to downstream objectives. Instruction fine-tuning plays a pivotal role in this process by training models on structured instruction–response pairs that explicitly define task goals, output formats, and expected reasoning behavior. Prior work has shown that instruction tuning substantially enhances reasoning performance, output consistency, and alignment with decision-oriented tasks (Vaillancourt & Thompson, 2024).

In the context of scam risk assessment, such alignment is crucial for producing stable, machine-parseable outputs, including numerical risk scores, categorical safety levels, and explanatory rationales, while maintaining consistent behavior across multi-turn conversational inputs. The development of data-driven approaches for fraud and scam detection is fundamentally limited by the scarcity of high-quality, real-world conversational data. Privacy regulations, ethical considerations, and reporting bias severely constrain access to authentic scam dialogues, particularly those involving vulnerable populations such as older adults. As a result, synthetic data generation has emerged as a practical and scalable alternative for modeling complex conversational behaviors in a controlled and privacy-preserving manner. Synthetic conversational data are typically generated using language models guided by predefined scenario templates and structural constraints. Existing studies demonstrate that such data can effectively capture grounded dialogue structures, interaction dynamics, and contextual dependencies relevant to downstream learning tasks (Abdullin et al., 2023; Bao et al., 2023). Within fraud modeling, synthetic dialogues enable systematic coverage of diverse scam strategies, escalation trajectories, and communication platforms that are rarely balanced or consistently represented in real-world datasets.

Importantly, the utility of synthetic data does not depend on faithfully replicating real conversations, but rather on exposing models to representative linguistic cues and decision-relevant patterns. Empirical evidence suggests that synthetic data can improve model efficiency, generalization, and robustness, particularly in low-resource or privacy-constrained settings (Gholami & Omar, 2023; Maheshwari et al., 2024). When combined with instruction fine-tuning, synthetic datasets provide structured supervision that supports stable learning of task-specific behaviors, even when employing compact model architectures.

For fraud detection tasks in which risk signals emerge gradually across conversational turns, synthetic multi-turn dialogues are particularly valuable. They enable explicit modeling of temporal escalation, uncertainty resolution, and behavioral shifts over time—phenomena that are difficult to capture using static or sparsely annotated real data. Prior applications of language models to fraud detection further suggest that carefully designed synthetic data can serve as a strong foundation for scalable, privacy-aware, and real-time risk assessment systems (Malingu et al., 2025). Evaluating models trained on such incremental and temporally evolving data requires error measures that reflect both numerical deviation and semantic proximity on a continuous risk scale. In progressive scam detection, risk does not shift abruptly between discrete classes, but instead accumulates as conversational evidence unfolds over time. Consequently, evaluation metrics must capture the magnitude of over- or under-estimation across turns rather than relying solely on rigid classification accuracy.

To quantify risk-score prediction accuracy, we adopt the Mean Absolute Error (MAE):

$$MAE = (1/N)\ \Sigma\ |\hat{y}_i - y_i| \qquad (1)$$

where $y_i \in [0–10]$ is the ground-truth risk score and $\hat{y}_i$ denotes the model prediction for the i-th cumulative dialogue instance. MAE provides an interpretable measure in the same units as the risk scale, making it well suited for evaluating gradual deviations in incremental, turn-based assessment.

## III. Methodology

We formulate elderly scam detection as progressive risk inference over cumulative conversational states. Rather than treating dialogue as a static classification input, our approach models scam risk as a temporally evolving signal that must be estimated under partial and expanding context. The proposed framework in Fig. 1 consists of three tightly coupled components: a synthetic multi-turn data construction strategy, a unified continuous risk scoring scheme, and parameter-efficient instruction tuning of small language models for structured risk estimation.

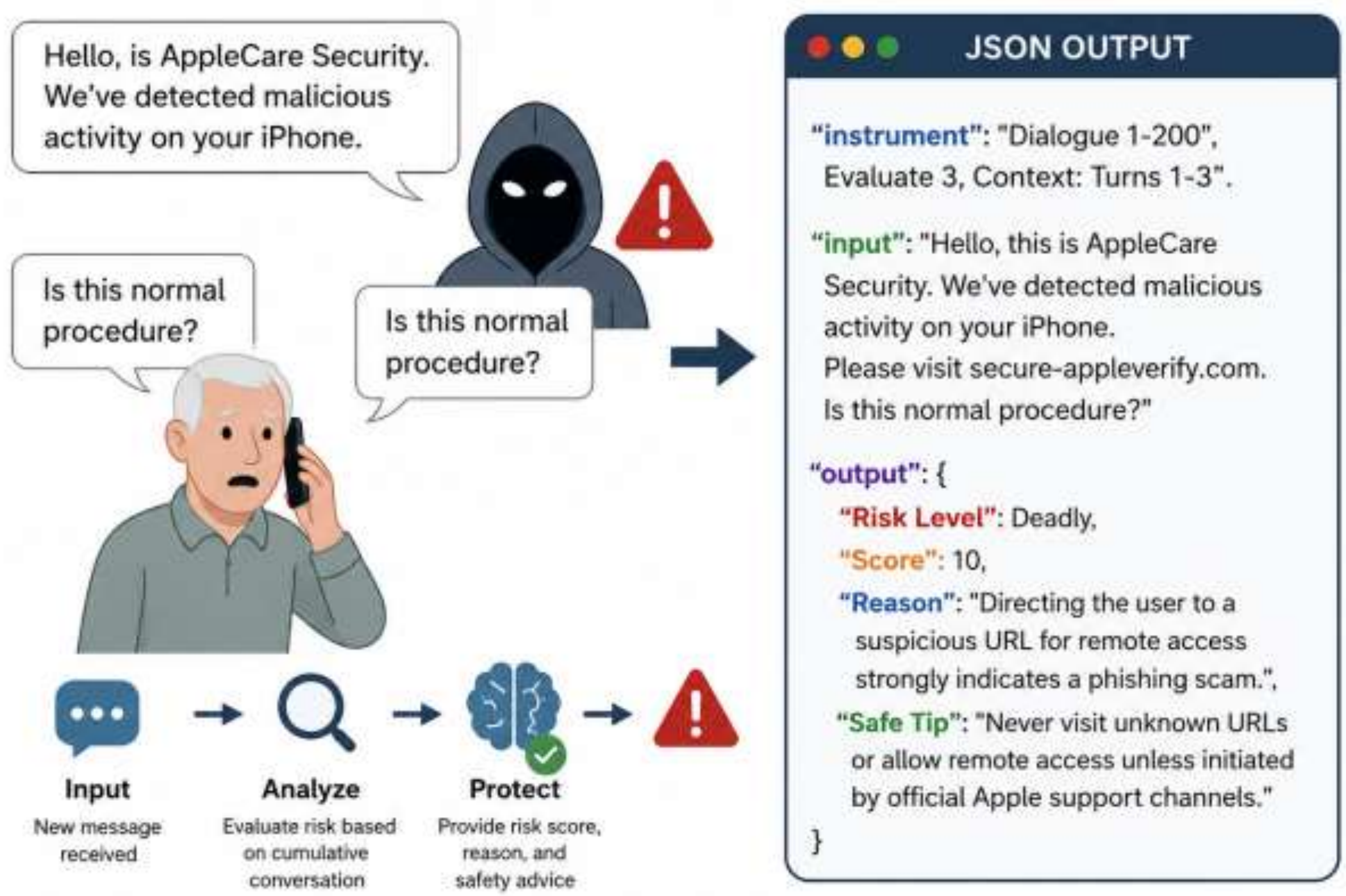

*Fig. 1. From conversational input to structured scam risk output.*

### *A. Synthetic Data Collection and Annotation Pipeline: Cumulative Turn-Based Data Preparation*

In this study, we formulate elderly financial scam detection as an incremental conversational risk inference problem over multi-turn interactions. Due to the scarcity and privacy sensitivity of real-world scam dialogues, we construct a controlled synthetic dataset that approximates realistic impersonation scenarios while enabling systematic supervision. Dialogue instances are generated using ChatGPT-4o and structured as two-party interactions between a scammer and an older adult victim. Each dialogue contains between two and eight turns, where a turn is defined as a paired exchange consisting of one message and its corresponding reply. This paired turn formulation is adopted as a deliberate modeling unit to preserve interaction coherence and to capture the temporal progression of scam behavior, including trust establishment, emotional manipulation, escalation strategies, and eventual requests for sensitive information or financial action. The final dataset comprises 600 training dialogues and 40 held-out test dialogues.

The dataset is constructed using a structured and balanced design across two primary scam domains: investment and charity fraud, and tech support fraud. Within each domain, approximately one hundred distinct scenarios are systematically defined and distributed across multiple fine-grained sub-categories to ensure coverage of diverse fraud strategies and interaction patterns. Dialogues are instantiated across common communication channels, including email, SMS, and phone calls, enabling variation in tone, urgency cues, and linguistic style.

The investment and charity domain captures scenarios involving financial account updates, performance evaluations, balance notifications, and transaction confirmations. The tech support domain models security alerts, system compromise claims, and device protection warnings. Across both domains, paired legitimate and scam versions are generated under comparable structural conditions to minimize superficial lexical cues and encourage models to learn intent- and behavior-level distinctions rather than keyword-based shortcuts. For each sub-category and communication channel, we generate paired dialogue instances in legitimate and scam variants under matched surface conditions. The two variants are designed to exhibit similar topical content and linguistic style, while differing in underlying intent, escalation trajectory, and the presence of action-oriented requests. This pairing functions as a controlled contrast set, improving class balance and reducing the likelihood that models succeed via keyword heuristics rather than behavior- and intent-level cues.

We perform annotation at the turn level to support incremental, real-time monitoring. Each turn is labeled independently using a separate large language model as an automated rater, producing three supervision signals: a binary scam indicator, an ordinal risk category, and a continuous risk score (0–10) reflecting the strength of fraud evidence. A subset of annotated turns was manually reviewed to verify the plausibility and consistency of assigned risk labels, numerical scores, and conversational escalation patterns, as well as to identify potential systematic errors in the automated annotation process. The review focused on detecting unrealistic risk transitions, inconsistent scoring behavior, and formatting irregularities in structured outputs. Importantly, dialogue generation and risk annotation were performed using separate language models to reduce model-specific bias and mitigate potential information leakage between data synthesis and supervision.

Instead of treating an entire dialogue as a single static classification unit, we reformulate scam detection as progressive risk inference over cumulative conversational context. Training samples are constructed by incrementally aggregating turns: the first instance contains only the initial exchange, and each subsequent instance appends the next turn, continuing until the full dialogue is observed. This transformation converts every multi-turn interaction into a sequence of temporally ordered, expanding contexts. Such cumulative representation reflects the operational reality of scam detection, where risk must be estimated under partial and evolving information. Early contexts are typically ambiguous and weakly informative, whereas later contexts reveal clearer escalation signals, behavioral pressure, or intent-driven requests. By exposing the model to this temporal progression, the framework encourages sensitivity to risk emergence, stabilization, and resolution patterns across conversational time. Each cumulative context is serialized as a structured input instance with dialogue identifiers and turn indices, paired with a standardized risk evaluation prompt. The target output comprises a categorical risk label, a continuous risk score, a brief explanatory rationale, and a user-facing safety recommendation. Dialogue generation and risk labeling are conducted using separate language models to reduce model-specific bias and minimize supervision leakage. This turn-aware formulation enables fine-grained analysis of early warning detection, delayed recognition, and uncertainty calibration in multi-turn scam interactions.

Although the proposed dataset is designed to capture realistic conversational escalation patterns, the dialogues remain synthetically generated and may not fully represent the linguistic diversity and behavioral complexity of real-world scam interactions. To mitigate potential generation bias, dialogue synthesis and risk annotation are performed using separate language models, and a subset of annotated instances is manually reviewed for plausibility and consistency. The primary objective of the synthetic framework is not exact replication of real conversations, but controlled exposure to representative scam-related behavioral and linguistic patterns under privacy-preserving conditions. Future work will extend the framework using real-world annotated conversational data.

### *B. Risk Class Definition and Numerical Scoring Scheme*

We represent scam risk using a continuous scale from 0 to 10, where 0 denotes fully benign interaction and 10 indicates explicit and imminent scam intent. This formulation moves beyond binary fraud classification by capturing degrees of suspicion and escalation intensity across conversational turns. In multi-turn settings, risk signals often emerge gradually rather than abruptly; a continuous scale therefore enables finer discrimination between early ambiguity, moderate concern, and high-confidence malicious intent. For interpretability and evaluation, the continuous scores are mapped to discrete semantic risk categories, allowing both regression-style estimation and class-based analysis. This hybrid design supports tolerance-aware evaluation, calibration assessment, and error stratification across varying levels of conversational risk. The correspondence between numerical ranges and semantic classes is summarized in Table I.

*TABLE I. MAPPING BETWEEN RISK CLASSES AND SCORE RANGES*

| Risk Class | Score Range | Interpretation |
|---|---|---|
| Safe | 0–1 | Benign content with no scam indicators |
| Benign | 2–3 | Neutral or informational content |
| Uncertain | 4 | Ambiguous signals, insufficient evidence |
| Suspicious | 5 | Early warning signs, mild suspicion |
| High Risk | 6 | Elevated concern, behavioral red flags |
| Severe | 7–8 | Strong scam indicators and pressure tactics |
| Critical | 9 | Imminent risk, clear malicious intent |
| Confirmed Scam | 10 | Explicit scam attempt requiring immediate action |

This scoring formulation enables evaluation beyond exact numerical matching. Because adjacent scores correspond to semantically similar risk states, we adopt a tolerance-aware perspective in which minor deviations (±1) are treated as near-consistent predictions, while larger deviations reflect substantive semantic disagreement. Such a design acknowledges the inherent uncertainty in early-stage conversational assessment, where risk boundaries are gradual rather than discrete. By integrating continuous estimation with category-level interpretation, the framework supports both regression-based calibration analysis and discrete safety-oriented evaluation. This dual representation is particularly important in real-world deployment, where systems must balance sensitivity to early warning signals with robustness against overestimation.

### C. Methods and Models

We fine-tune several publicly available small language models on the proposed cumulative turn-based dataset for three epochs. Training is performed on cumulative dialogue instances derived from 600 synthetic multi-turn training dialogues spanning approximately 200 scenario templates.

To analyze how different inductive biases affect conversational risk estimation, we select four representative architectures: Phi-4-mini-instruct, LLaMA-3.2-1B, Qwen3-1.7B, and DeepSeek-R1-Distill-Qwen-1.5B. These models represent instruction-aligned, dialogue-oriented, and reasoning-distilled paradigms within a comparable 1–4B parameter scale, enabling controlled comparison under similar capacity constraints. All experiments use our fine-tuned checkpoints, and implementation details are available upon request. Parameter-efficient adaptation is conducted using Low-Rank Adaptation (LoRA) (Hu et al., 2022) with PiSSA initialization (Meng et al., 2024), allowing task-specific updates without modifying the full model weights. After training, the adapter parameters are merged into the base models to produce standalone checkpoints for inference.

Within the selected model set, Qwen3-1.7B and DeepSeek-R1-Distill-Qwen-1.5B are included to provide complementary inductive biases for dialogue-based risk assessment. Qwen3-1.7B emphasizes conversational fluency and multi-turn interaction modeling, whereas DeepSeek-R1-Distill-Qwen-1.5B is distilled from a reasoning-oriented architecture and prioritizes multi-step analytical behavior. Comparing these two models allows us to examine how dialogue-centric versus reasoning-centric training objectives influence incremental risk estimation across conversational turns. Fig. 2 presents the overall end-to-end pipeline for cumulative conversational risk assessment using fine-tuned small language models.

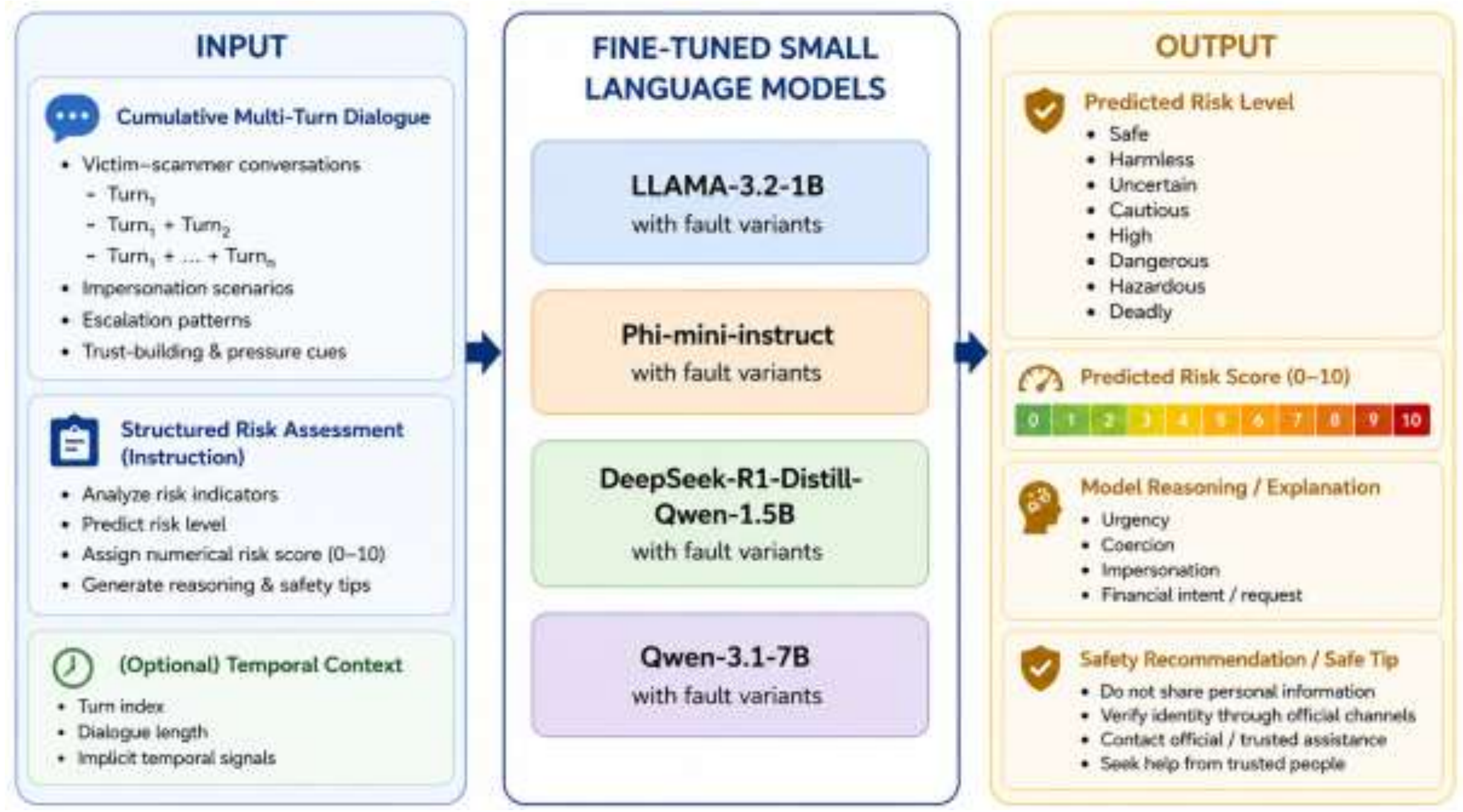

*Fig. 2. End-to-end pipeline for elder fraud risk assessment using fine-tuned small language models.*

## IV. Results and Model Comparison

Among the evaluated models, the fine-tuned LLaMA-3.2-1B achieves the most stable overall performance under cumulative risk estimation. It consistently generates valid, machine-parseable outputs and obtains the lowest Mean Absolute Error (MAE = 1.56) in risk-score prediction. Beyond numerical accuracy, the model maintains high compliance with the predefined structured output format and preserves the ordinal progression of risk levels across dialogue turns, indicating reliable calibration under incremental context accumulation. The fine-tuned Phi-4-mini-instruct model demonstrates moderate performance (MAE = 2.06). While its numerical accuracy remains competitive, the model exhibits reduced consistency in structured output adherence. Error analysis indicates a systematic positive bias in low-risk scenarios, where predicted scores tend to exceed ground-truth values. This overestimation behavior increases the frequency of large deviations and reflects a conservative calibration tendency under ambiguous early-stage contexts.

The fine-tuned Qwen-based models show comparatively lower robustness in structured risk estimation. The fine-tuned Qwen-1.5B model exhibits substantial numerical deviation (MAE = 5.58) and inconsistent format compliance, suggesting difficulty in jointly maintaining structured analytical output and calibrated scoring. The fine-tuned Qwen3-1.7B model frequently fails to transition from conversational continuation to structured evaluation mode, resulting in a high proportion of non-parseable outputs. Consequently, quantitative risk-score analysis is not applicable for this model due to insufficient valid predictions.

*TABLE II. PERFORMANCE COMPARISON OF FINE-TUNED SMALL LANGUAGE MODELS ON STRUCTURED SCAM RISK SCORING*

| Fine-Tuned Model | Base Model | Parameters | Architecture Origin | Instruction Strength | Output Quality | MAE (Risk Score) |
|---|---|---|---|---|---|---|
| fine-tuned-mini-instruct | microsoft/Phi-4-mini-instruct | 3.84B | Microsoft | Moderate | Inconsistent | 2.06 |
| fine-tuned-3.2-1B | meta-llama/LLaMA-3.2-1B | 1.3B | Meta AI | Strong | Consistent | 1.56 |
| fine-tuned-Qwen-1.5B | deepseek-ai/Qwen-1.5B-distill | 1.5B | DeepSeek | Weak | Broken or missing | 5.58 |
| fine-tuned-Qwen3-1.7B | Qwen/Qwen3-1.7B | 1.7B | Qwen | Very weak | Not Usable | N/A |

Fig. 3 visualizes the relationship between ground-truth and predicted risk scores for the fine-tuned Phi-4-mini-instruct model. The scatter distribution indicates a systematic positive deviation in low-risk regions (0–4), where predicted scores frequently exceed the corresponding ground-truth values. This pattern is consistent with the overestimation tendency identified in the tolerance-based analysis and reflects conservative calibration under early or ambiguous conversational contexts. Fig. 4 presents the corresponding analysis for the fine-tuned LLaMA-3.2-1B model. Predictions are more tightly concentrated around the diagonal, suggesting improved calibration and reduced variance across risk levels. Score transitions across dialogue turns exhibit smoother progression, indicating stronger preservation of ordinal structure under cumulative context expansion.

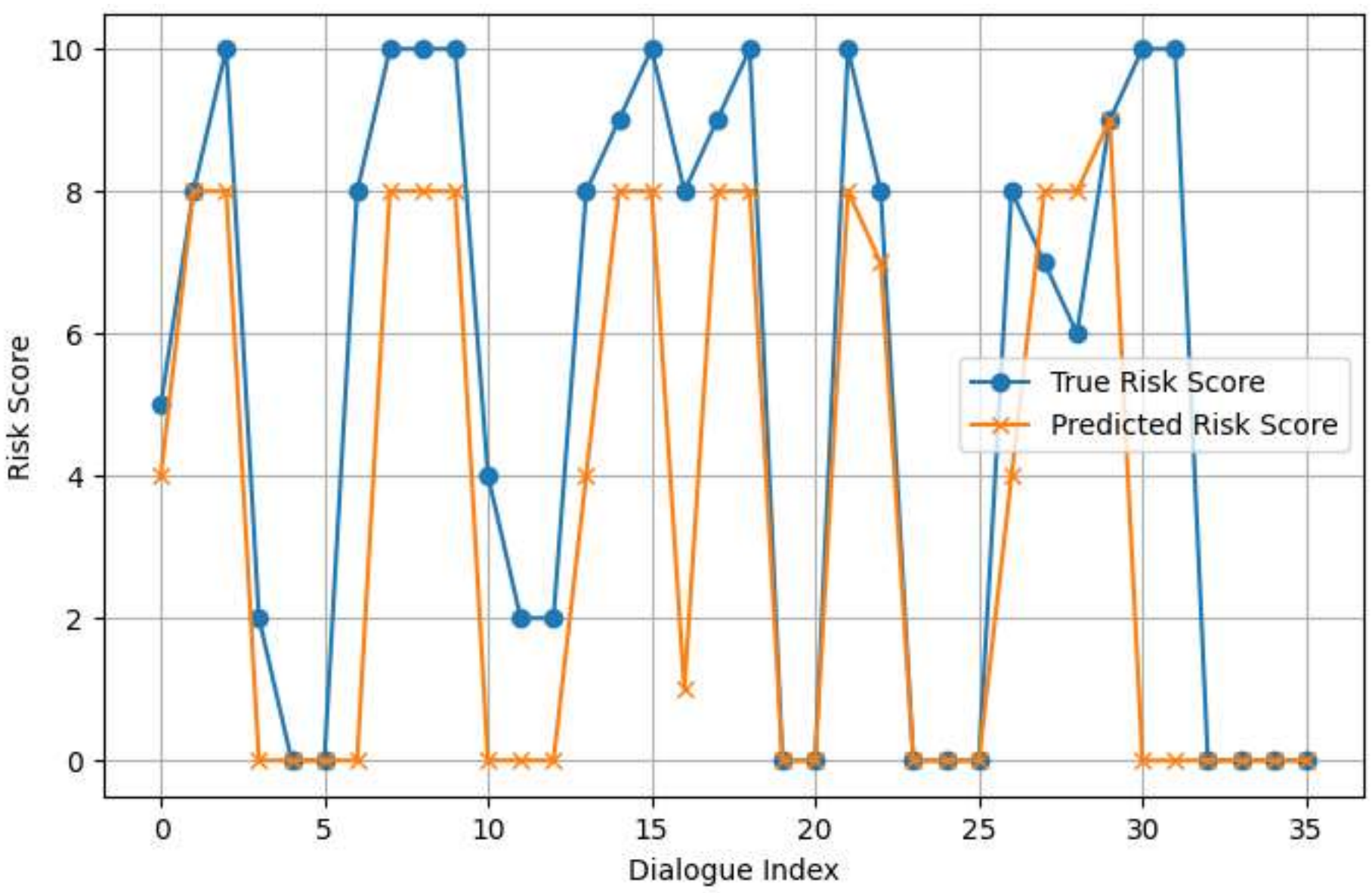

*Fig. 3. True vs. predicted risk scores for fine-tuned Phi-4-mini-instruct.*

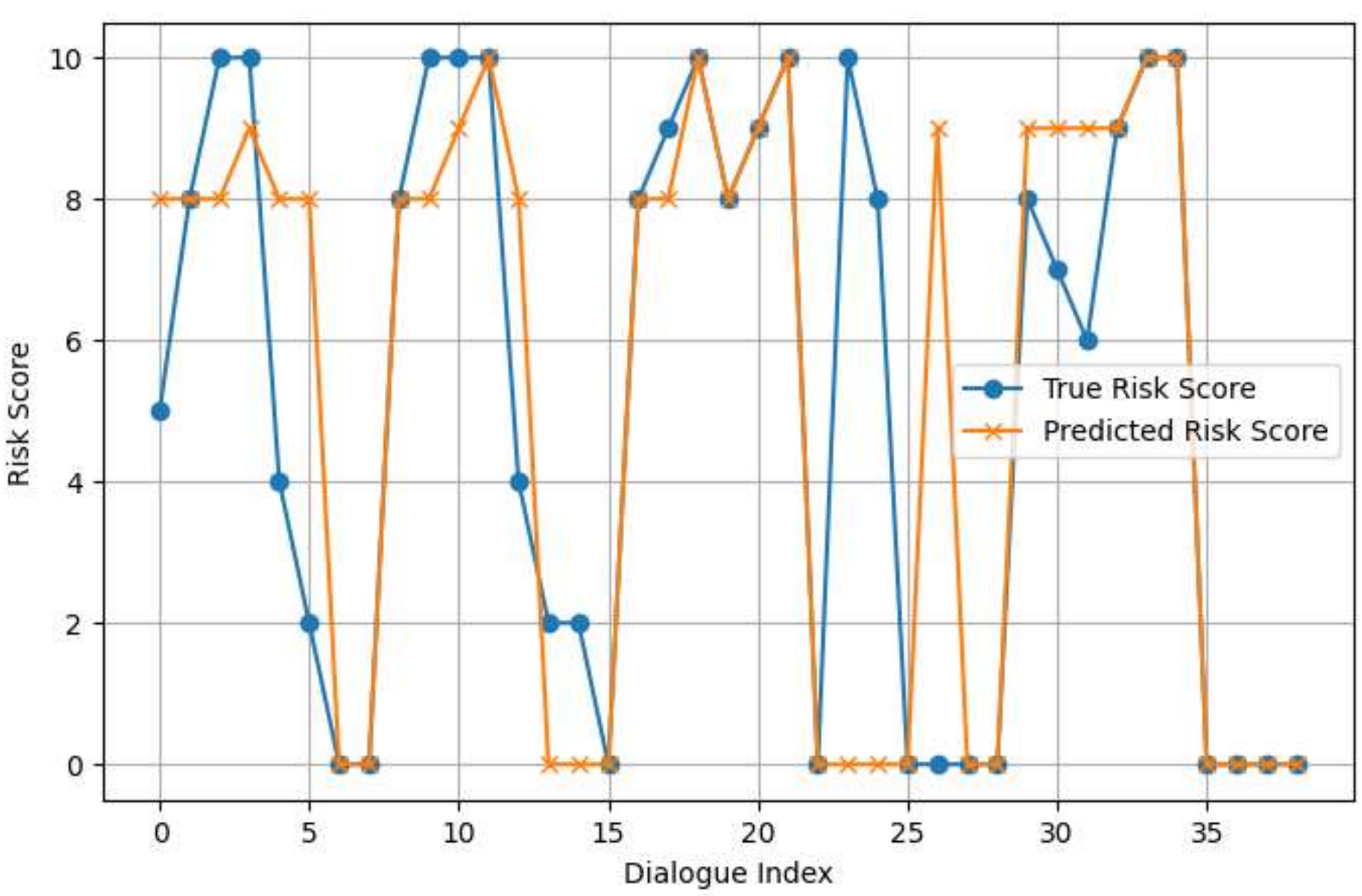

*Fig. 4. True vs. predicted risk scores for fine-tuned LLaMA-3.2-1B.*

To further analyze model behavior beyond aggregate error metrics, we conduct a tolerance-based deviation analysis that captures both the magnitude and direction of prediction errors. The fine-tuned Phi-4-mini-instruct model exhibits a consistent positive bias in low-risk regions (0–4), where predicted scores frequently exceed the corresponding ground-truth values. In several cases, benign or low-risk inputs are assigned high predicted scores (8–9), resulting in large upward deviations. This pattern indicates conservative calibration under ambiguous early-stage contexts, with a tendency to prioritize sensitivity over precision.

In contrast, the fine-tuned LLaMA-3.2-1B model demonstrates a more balanced deviation profile. Most prediction errors remain within a narrow range (typically ±1–2 risk-score points), and the ordinal progression of risk levels across dialogue turns is largely preserved. Larger deviations occur less frequently and are primarily associated with abrupt contextual shifts within the dialogue.

For the Phi-4-mini-instruct model, valid format-compliant outputs were obtained for 35 of the 40 evaluation cases. The remaining five instances were excluded from deviation analysis due to non-parseable or incomplete structured outputs.

*TABLE III. TOLERANCE-BASED ERROR DISTRIBUTION AND RISK ESTIMATION BIAS FOR FINE-TUNED MODELS*

| **Model** | **≤1 (OK)** | **=2** | **3–4** | **≥5** | **Total** |
|---|---|---|---|---|---|
| LLaMA-3.2-1B (FT) | 24 (60.0%) | 7 (17.5%) | 4 (10.0%) | 5 (12.5%) | 40 |
| Phi-4-mini (FT) | 17 (48.6%) | 9 (25.7%) | 3 (8.6%) | 6 (17.1%) | 35 |

## V. Conclusion and Future Work

This study formulates elderly fraud detection as incremental risk estimation over cumulative conversational context. By combining structured synthetic data with instruction-tuned small language models, the framework enables temporally aware and machine-parseable risk assessment rather than static classification. Experimental results show that instruction adherence and output-format stability are more critical than parameter scale alone for reliable multi-turn risk scoring.

### *A. Limitations and Real-World Deployment*

Although the proposed framework demonstrates promising performance under controlled synthetic settings, several limitations remain. The current study focuses primarily on text-based conversational interactions and does not incorporate multimodal signals such as voice characteristics, emotional cues, or behavioral metadata that may further improve scam detection reliability. In addition, synthetic dialogues may not fully capture the linguistic diversity, unpredictability, and evolving strategies observed in real-world scam behavior across different communication platforms. The proposed cumulative turn-based formulation also assumes relatively coherent conversational progression, whereas real-world interactions may contain interruptions, topic shifts, incomplete exchanges, or adversarial manipulation attempts. Furthermore, although compact language models provide practical deployment potential for resource-constrained environments, real-world deployment still requires careful consideration of privacy preservation, continual adaptation, and robustness under dynamically evolving fraud patterns.

Future work will evaluate generalization on real-world annotated data, extend the framework to multilingual settings, investigate multimodal conversational analysis, and explore privacy-preserving deployment strategies such as on-device inference and federated adaptation.